\documentclass[runningheads]{llncs}

\usepackage[mobile]{eccv}
\usepackage{threeparttable}
\usepackage{caption}
\usepackage{subcaption}
\usepackage{tabularray}
\usepackage{blindtext}

\usepackage{eccvabbrv}

\usepackage{graphicx}
\usepackage{booktabs}

\usepackage[accsupp]{axessibility}  % Improves PDF readability for those with disabilities.

\usepackage[pagebackref,breaklinks,colorlinks,citecolor=eccvblue]{hyperref}
\usepackage{amsmath,amsfonts,bm}

\def\eqref#1{equation~\ref{#1}}
\def\1{\bm{1}}

\def\vf{{\bm{f}}}

\def\vx{{\bm{x}}}

\def\vz{{\bm{z}}}

\def\mI{{\bm{I}}}

\DeclareMathAlphabet{\mathsfit}{\encodingdefault}{\sfdefault}{m}{sl}
\SetMathAlphabet{\mathsfit}{bold}{\encodingdefault}{\sfdefault}{bx}{n}

\usepackage{colortbl}
\usepackage[ruled,vlined]{algorithm2e}
\usepackage{orcidlink}

\begin{document}

\title{TexRO: Generating Delicate Textures of 3D Models by Recursive Optimization} 
% \title{Supplementary material for ``TexRO: Generating Delicate Textures of 3D Models by Recursive Optimization''}

\titlerunning{TexRO}

\author{Jinbo Wu \and
Xing Liu \and
Chenming Wu \and
Xiaobo Gao \and
Jialun Liu \and
Xinqi Liu \and  \\
Chen Zhao \and
Haocheng Feng \and
Errui Ding  \and
Jingdong Wang
}
% Second Author\inst{2,3}\orcidlink{1111-2222-3333-4444} \and
% Third Author\inst{3}\orcidlink{2222--3333-4444-5555}}

\authorrunning{J. Wu et al.}

\institute{Department of Computer Vision Technology (VIS), Baidu Inc.}
% Princeton University, Princeton NJ 08544, USA \and
% Springer Heidelberg, Tiergartenstr.~17, 69121 Heidelberg, Germany

\maketitle

% \begin{abstract}
%   The abstract should concisely summarize the contents of the paper. 
%   While there is no fixed length restriction for the abstract, it is recommended to limit your abstract to approximately 150 words.
%   Please include keywords as in the example below. 
%   This is required for papers in LNCS proceedings.
%   \keywords{First keyword \and Second keyword \and Third keyword}
% \end{abstract}

% \begin{abstract}
% We introduce TexRO, an innovative solution to synthesize intricate textures on 3D models via view optimization. Our technique addresses the challenge of view-dependent texture synthesis, ensuring coherency across diverse viewing angles through a unique multiview consistency loss function.
% TexRO incorporates diffusion networks, facilitating synthesis of complex textures while preserving subtle details. The model optimizes texture generation by effectively managing view-dependent effects, demonstrating improvement over existing methods.
% Comprehensive experiments confirm TexRO's superiority in texture quality, detail preservation, and visual consistency over contemporary methods. Its versatility is validated through successful application on diverse 3D models. The source codes will be provided to the public soon.
% \end{abstract}% 

\begin{abstract}
This paper presents TexRO, a novel method for generating delicate textures of a known 3D mesh by optimizing its UV texture. The key contributions are two-fold. We propose an optimal viewpoint selection strategy, that finds the most miniature set of viewpoints covering all the faces of a mesh. Our viewpoint selection strategy guarantees the completeness of a generated result. We propose a recursive optimization pipeline that optimizes a UV texture at increasing resolutions, with an adaptive denoising method that re-uses existing textures for new texture generation. Through extensive experimentation, we demonstrate the superior performance of TexRO in terms of texture quality, detail preservation, visual consistency, and, notably runtime speed, outperforming other current methods. The broad applicability of TexRO is further confirmed through its successful use on diverse 3D models. Project page: \href{https://3d-aigc.github.io/TexRO}{https://3d-aigc.github.io/TexRO}.

% \begin{abstract}
% With the evolution of diffusion probabilistic models, 3D textures can be synthesized by employing pre-trained 2D diffusion models with 3D-aware guidance, such as depth derived from geometry. Recent findings have shown that simultaneously generating multiple views helps diffusion models produce textures that align with the underlying geometry. 
% To improve texture generation quality, this paper introduces TexRO, a novel approach that fabricates realistic textures on 3D geometries using multiview optimization. The key insight guiding us is that effective view selection and recursive multi-view texture optimization from the initial state are crucial for producing high-quality, realistic textures. TexRO completes generation directly on texture maps, thus circumventing potential losses in precision and fidelity that often occur with extra post-processing steps.
% Through extensive experimentation, we demonstrate the superior performance of TexRO in terms of texture quality, detail preservation, visual consistency, and, notably runtime speed, outperforming other current methods. The broad applicability of TexRO is further confirmed through its successful use on diverse 3D models. 

% Project page: \href{https://tex-opt.github.io}{https://tex-opt.github.io}.
\end{abstract}
% \xing{To be revised later.}    
\begin{figure}[h]
    \centering
    \captionsetup{type=figure}
        \includegraphics[width=1.0\linewidth]{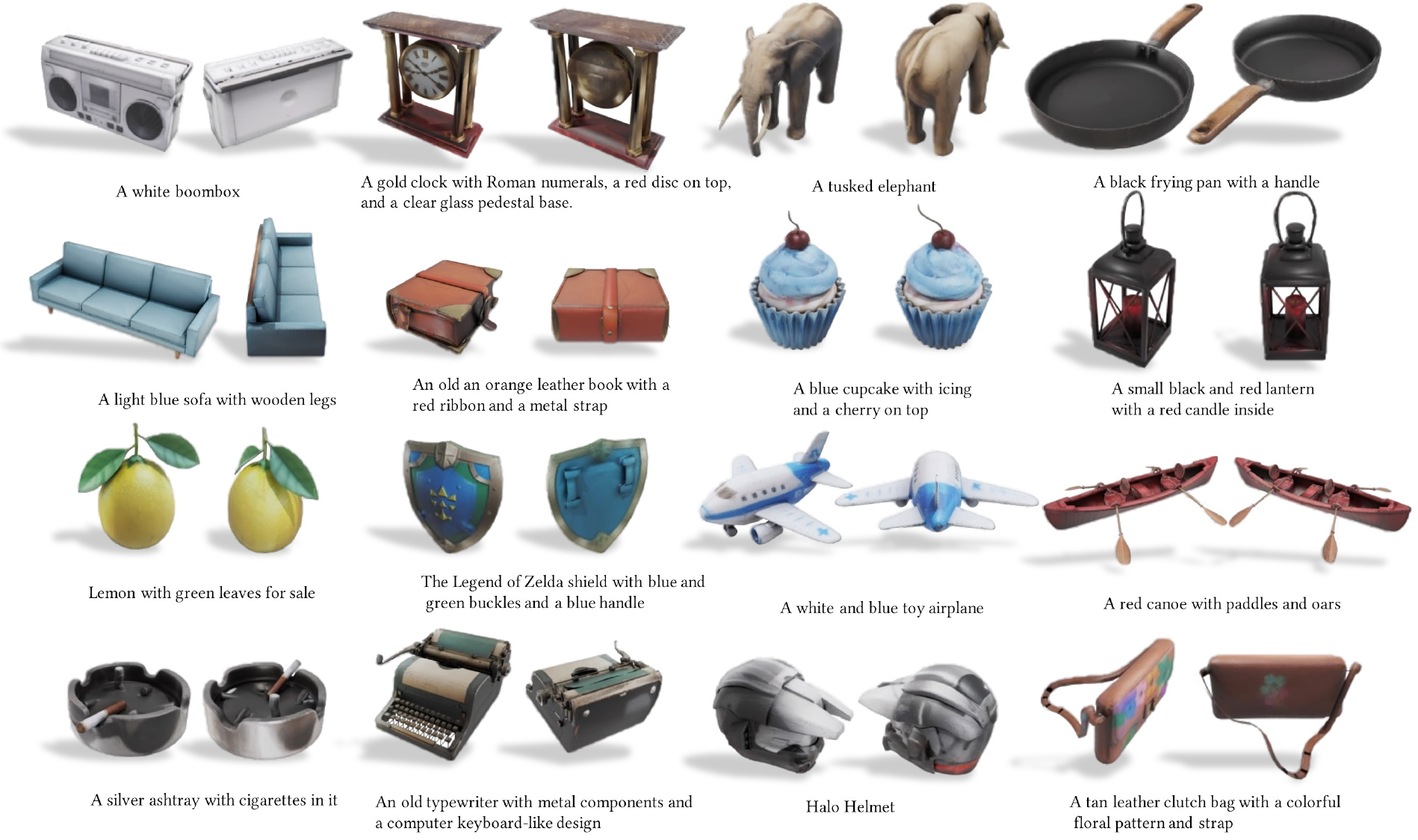}
    \captionof{figure}{The proposed TexRO generates realistic textures for a known 3D mesh based on prompts. Our key contributions include: 1) a novel recursive optimization method that refines UV textures at increasing resolutions using the proposed interlaced denoising module, and 2) an effective viewpoints selection strategy. The proposed TexRO has achieved the fastest texture generation ($\sim 1$ min.) and the highest generation quality (as measured by FID and KID scores) compared to previous studies. Example results are showcased in the figure.} 
    \label{fig:figInconsistent}
\end{figure}

\section{Introduction}%\xing{I am shitting on this.}
\label{sec:intro}

3D content creation is a vital component in various applications, such as generating visual content for movies, games, and the AR/VR industry. Generative static 3D content creation encompasses two crucial domains of research: the development of high-fidelity geometries and the creation of visually captivating textures. Both tasks share a common objective of democratizing 3D content creation. In recent years, significant advancements have been made in 3D geometry generation, leading to remarkable enhancements in high-quality and diversity~\cite{zheng2023lasdiffusion,sdfusion,hong2023lrm,li2023instant3d}. Conversely, the controllable creation of detailed and delicate textures for 3D models remains a challenging area~\cite{chen2022auv, texturepaper, chen2023text2tex, cao2023texfusion}. 
% \chenming{Ref several works}. 

% This task can be divided into two topics to study, that of creating geometries with accurate surfaces and that of generating varied visually pleasing appearances characterized by high-quality textures. The first topic is comparably easy to succeed in the scenario of real applications, as there are effective approaches to creating geometries in the digital world. Photogrammetric approaches~\cite{ba, colmap} $have been developed for this purpose for a long time; In recent years, NeRF-based approaches \cite{neus, volsdf, nero} $have improved the performance of 3D reconstruction; More recent advancements \cite{sdfusion, zheng2023lasdiffusion} $have shown that 3D meshes with high-quality surfaces can be generated from pure text or image. 

Recently, diffusion models have made remarkable strides in image generation, largely due to training with abundant data from the Internet. Naturally, one might consider utilizing a 3D diffusion model, trained on 3D textured data, to generate textures (and shapes) within a volumetric 3D space, mirroring the operation of a 2D diffusion model.  However, this leads to less effective learning and often produces blurry outcomes~\cite{oechsle2019texture, metzer2023latent}. Moreover, regardless of the optimization using efficient latent representations \cite{eg3d}, training a 3D diffusion model to a satisfactory level requires a significant amount of 3D data and extensive computational resources. An alternative approach involves using distillation sampling to optimize 3D fields~\cite{poole2022dreamfusion, wang2023prolificdreamer}. However, these methods often bear the disadvantage of lengthy optimization time, ranging from tens of minutes to hours. 

In contrast, we introduce an efficient approach called TexRO for generating realistic textures for a given 3D model. The proposed approach has two stages. 
% through recursive optimization. 
% This involves leveraging a 
% pre-trained 2D diffusion model to produce multiple images from a set of optimized viewpoints, which are then projected onto the mesh surface to create an initial texture. The texture is then iteratively refined through recursive optimization. 
% The advantages of our proposed method are twofold: firstly, it offers faster computation compared to methods mentioned above; secondly, it harnesses the strengths of well-established 2D diffusion models. 
In the first stage, we generate an initial UV texture using a readily available depth-driven 2D diffusion method~\cite{shi2023zero123plus}. This initialization process mitigates the issues posed by other 2D diffusion models that fail to retain multi-view consistency. In addition, we propose an optimal viewpoint selection strategy that finds the most miniature set of views covering all the faces of a fmesh. This strategy ensures the completeness of the generated textures. In the second stage,  
% Taking advantage of the presence of coarse textures, we can render the 3D model from those selected viewpoints using differential rendering. However, individually optimizing each rendered view by the 2D diffusion model results in severe multi-view inconsistency. To tackle this issue, 
we propose a recursive optimization scheme, where an ``adaptive denoising'' strategy is employed to optimize the textures. This strategy is capable of re-using existing textures to generate new textures by adaptively injecting noises into pixels with different schedulers. Overall, the recursive optimization scheme optimizes a UV texture at increasing resolutions in RGB space. This design is proposed based on the insight that optimizing a UV texture at a low resolution facilitates generating consistent content while doing so at a high resolution produces details. Our proposed approach is empirically validated to be robust against changes in rendering view, including alterations in position and orientation.
% The image of the current view is retained by aligning the noise schedulers and performing the adaptive denoising operation over the overlapped regions between different viewpoints.
% We propose a multi-resolution UV texture optimization to ensure consistent texture generation while maintaining sufficient detail in the generated view images.
% Our proposed adaptive diffusion processes exhibit resilience against changes in render view, such as alterations in positions and orientations. 

% adaptive processes\chenming{what?} within the RGB space, detailed textures are constructed, particularly under the viewpoints selected earlier\chenming{why?}. While our method draws inspiration from~\cite{cao2023texfusion}, which performs texture generation in the latent space, it distinguishes itself through its ability to eliminate errors typically associated with decoding low spatial resolution latent features via rasterization.

% As a result, our method necessitates no extra processing\chenming{extra processing is decoding? It has no relationship with "confined to the UV"} to obtain the textured mesh, as its operations are confined exclusively to the UV space. 

% \chenming{Not mentioned recursive at all.}
We conduct extensive experiments on widely-used 3D datasets to validate the effectiveness and efficiency of our proposed method. The results of our experiments manifest that our method can generate textures that maintain global coherence and exhibit intricate details. These textures align well with the conditioning text prompts, as exemplified in Fig~\ref{fig:figInconsistent}. The quantitative data indicate that our method significantly advances texture generation for 3D models. Additionally, the qualitative results further corroborate this conclusion. The key contributions of our work can be summarized as follows:
\begin{itemize}
    \item We introduce \emph{TexRO}, a recursive optimization-based approach, combined with an adaptive denoising method for faster, realistic texture synthesis on 3D geometries leveraging the advancements of pre-trained 2D diffusion models.
    \item We propose an optimal viewpoint selection strategy to confine an effective optimization space for our proposed recursive optimization.
    \item Comprehensive experiments and user studies are undertaken to demonstrate our method surpasses existing approaches and represents the cutting edge of current capabilities.
\end{itemize}

\section{Related Work}

% \subsection{Image-based Texturing}
% The technique of texturing a mesh from reconstruction by multi-view stereopsis has been studied for several decades. 
% Mvs-Texturing \cite{waechter2014let}

\subsection{Texture Synthesis}
\noindent \textbf{Image-based Texture Synthesis}~has been studied for several decades. 
Waechter et al.~\cite{waechter2014let}
present a comprehensive framework for automatic texturing of large-scale 3D reconstructions from images. It addresses challenges like varying image properties, occluders, and geometry inaccuracies. A graph-cut labeling approach selects optimal textures, followed by global and local color adjustment steps. 
Zhou et al.~\cite{zhou2014color} addresses the problem of creating colored 3D models using consumer depth cameras, an energy function based on photo-consistency, smoothness, and visibility, is optimized efficiently using belief propagation.
Fu et al.~\cite{fu2018texture} first selects an optimal image for each model face, and then performs global camera pose optimization and local texture coordinate refinement to align textures across faces.
Huang et al.~\cite{huang2020adversarial} learns a discriminator as a patch-based misalignment-tolerant metric to guide texture optimization and make it robust to errors in geometry and camera poses.
Thies et al.~\cite{thies2019deferred} interprets high-dimensional neural textures, which store more information than traditional textures and are capable of handling imperfect 3D content.
Yariv et al.~\cite{yariv2023bakedsdf} and Tang et al.~\cite{tang2023delicate} optimize a hybrid neural volume-surface model designed for surface reconstruction, then extract a mesh and an appearance texture for real-time rendering.
% \chenming{Add recent works such as Paint3D}

\vspace{4pt}
\noindent \textbf{Example-based Texture Synthesis}~primarily involves applying exemplary patterns across a surface, frequently employing a distinct direction field to dictate local orientation, has been well-studied in the computer graphics community. Example-based texture synthesis enables easy content creation and shows promise for bridging procedural and by-example texturing~\cite{wei2009state}, such as \cite{lefebvre2006appearance} transforms a texture exemplar into an appearance space before synthesis to improve quality and enable new functionalities, \cite{kopf2007solid} extends 2D texture optimization methods to 3D while integrating global histogram matching to improve synthesis quality and convergence.
Gatys et al.~\cite{gatys2015texture} utilize image latent features extracted by convolutional networks to synthesize textures. GANs have also been explored in~\cite{zhou2018nonstationary, portenier2020gramgan} to generate more realistic example-based texture outputs.
Recently,
NeRF-Texture~\cite{huang2023nerf} leverages a texture generation network along with the radiance field to synthesize high-quality textures, overcoming the limitations of NeRFs in modeling complex textures.

\subsection{Learning to Generate Textures}

There exist several works that generate textures in 3D space or field. 
TexutreField~\cite{oechsle2019texture} represents texture as a continuous 3D function parameterized by a neural network. Any 3D point can be mapped to an RGB color value, and this process can be integrated with various shape representations, such as voxels, point clouds, or meshes.
EG3D~\cite{eg3d} introduces a novel hybrid explicit-implicit network architecture that improves computational efficiency and image quality without relying heavily on approximations.
RODIN~\cite{rodin} addresses the computational and memory-intensive nature of high-quality 3D avatar generation by projecting multiple 2D feature maps onto a single plane for efficient 3D-aware diffusion. 
3DGen~\cite{gupta20233dgen} comprises a Variational Autoencoder (VAE) that encodes a colored point cloud into a triplane Gaussian latent space and reconstructs the textured mesh from this latent space and a diffusion model that generates the triplane features.
Dundar et al.~\cite{dundar2023fine} design a generative adversarial network in UV space to add realistic fine details to texture maps initialized by differentiable rendering, with an attention mechanism in the generator and a learnable position embedding in the discriminator.
Sin3DM~\cite{wu2023sin3dm} proposes a novel diffusion model that learns from a single 3D textured shape to generate high-quality variations with detailed geometry and texture using a similar triplane representation.
Point-UV Diffusion~\cite{yu2023texture} employs a denoising diffusion model and UV mapping to generate 3D consistent textures, where the semantics of a UV parameterization is an important factor towards high-quality results. Our work leverages a potent pre-trained 2D diffusion model, absent in the 3D realm, and optimizes directly within the color space, as opposed to a 3D space or field, to yield more captivating outcomes in rendered views.

A body of research~\cite{pavllo2020convolutional,xiang2021neutex,bhattad2021view,henderson20cvpr,tulsiani2020implicit,vmr2020,dibr,pan2020gan2shape,wu2020unsupervised,chen2022auv} studies texture generation on a template or aligned space, those methods are prone to lose details or fail to capture the shapes with complex structure or topology.
% AUV-Net~\cite{chen2022auv} learns to embed 3D surfaces into a 2D aligned UV space, thereby aligning textures across different 3D shapes and enabling easy synthesis of textures by generative image models. 
To enable texture generation for more general 3D models,
Texturify~\cite{siddiqui2022texturify} uses differentiable rendering and adversarial losses to learn high-quality surface texturing from real-world images.
% CLIP-Mesh~\cite{mohammad2022clip}
Latent-NeRF \cite{metzer2023latent} propagates gradients from a pre-trained 2D diffusion model through differentiable rendering to optimize the latent texture map.
TexFusion~\cite{cao2023texfusion} leverages latent diffusion models, applies a denoiser on a set of 2D renders of the 3D object, and aggregates the denoising predictions on a shared latent texture map.
TEXTure~\cite{texturepaper} and Paint3D \cite{zeng2023paint3d} also uses a pre-trained depth-to-image diffusion model to paint a 3D model from different viewpoints, while addressing inconsistencies through a trimap partitioning of the rendered image. 
Text2Tex~\cite{chen2023text2tex} also uses a pre-trained depth-aware image diffusion model to synthesize high-resolution partial textures from multiple viewpoints progressively. To avoid inconsistencies and artifacts, the authors introduce a generation mask and an automatic view sequence generation scheme. Distinct from prior research, our work proposes a multiview optimization scheme, employing view optimization to limit the optimizing space. This approach bolsters the quality and multiview consistency of the textures generated.
\section{Preliminaries and Problem Definition}
\noindent \textbf{Preliminary: Image Diffusion Models.}~~
A forward diffusion process is defined as gradually adding noises to an image until it becomes noisy. A diffusion model learns the reversed process for each step. Consequently, the model learns the underlying data distribution, enabling it to generate images. Let $p_{\theta}(\vx)$ denote the data distribution and $\vx_{0}$ represent a data sample. The transformation process models $\vx_{0}$ as a noisy version of a Gaussian random variable $x_{T}$, which can be formulated as $\vx_{t} = \sqrt{1-\alpha_{t}} \vx_{t-1} + \sqrt{\alpha_{t}} \boldsymbol{\epsilon}_{t}$, where $\boldsymbol{\epsilon}_{t} \sim \mathcal{N}(0, \mI)$ and $0 \leq \alpha_{t} \leq 1$ controls the noise level \cite{ho2020denoising}. 
This process is governed by a stochastic differentiable equation (SDE) that can be discretely approximated over time steps $t = 0, 1, ..., T$ \cite{song2019generative}. To generate new samples, we reverse these steps, where the transition probabilities $q_{\theta}(x_{t-1} | x_{t})$ are guided by a learned reverse-time SDE, by estimating the noise to remove as $\boldsymbol{\epsilon}_\theta^{(t_i)}$ using network $\theta$, allowing the model to gradually add structure to the noise distribution, resulting in samples from $p_{\theta}(x)$. 
We leverage iterative diffusion model sampling with Denoising Diffusion Implicit Models (DDIM)~\cite{ddim}:
\begin{equation}
\begin{aligned}
    \vx_{i-1} =& \sqrt{\alpha_{i-1}}\left(\frac{\vx_i - \sqrt{1 - \alpha_i} \boldsymbol{\epsilon}_\theta^{(t_i)}(\vx_i)}{\sqrt{\alpha_i}}\right) \\
    &+ \sqrt{1 - \alpha_{i-1} - \sigma_{t_i}^2} \cdot \boldsymbol{\epsilon}_\theta^{({t_i})}(\vx_i) + \sigma_{t_i} \boldsymbol{\epsilon}_{t_i},
\end{aligned}\label{eqn:ddim}
\end{equation}
where $\sigma_{t_i}$ is a variance parameter related to $\alpha_i$ and $\alpha_{i-1}$. For simplification, we denote $\vf_\theta^{(t_i)}(\vx_{i-1} | \vx_i)$ as the function to obtain $\vx_{i-1}$ from $\vx_i$ by using DDIM sampling.

\vspace{2pt} \noindent \textbf{Preliminary: Mesh and Differentiable Rasterization.}~~
In the realm of 3D modeling, it is customary to represent objects using discrete polygons. In our research, we utilize the popular triangular mesh, which consists of arranged triangles so that intersecting triangles only share an edge or a vertex. For a triangular mesh $\mathcal{M}=(\mathcal{V}, \mathcal{T})$ consisting of $N$ vertices and $M$ faces, we typically denote the set of vertices as $\mathcal{V} = \{v_1, v_2, \ldots, v_N\}$, where each $v_i$ represents the $i$-th vertex in 3D space. The triangles are denoted as $\mathcal{T} = \{t_1, t_2, \ldots, t_M\}$, where each $t_l$ is defined by three vertices forming a triangle.
Differentiable rasterization is primarily utilized to generate pixel-level derivatives for shading computations in graphics processing units at a given camera viewpoint $c$. Each vertex $v_i$ in the mesh $\mathcal{M}$ is attributed with distinct properties such as position, color, and texture coordinates in the UV space, where the UV space $(u, v)$ is a parameter space that maps a 2D texture $I$ onto 3D models. Each triangle $t_l$ is subsequently rasterized into a set of fragments $\mathcal{F} = \{f_1, f_2, \ldots, f_p\}$ on the imagery plane of camera $c$, we denote this process as $\mathcal{R}(\cdot)$ and its reverse process as $\hat{\mathcal{R}}(\cdot)$. 
For each fragment $f_j$, the attributes at that point are interpolated from the attributes of the vertices of the triangle $t_l$ that the fragment belongs to, including the UV coordinates, which are used to look up the corresponding texture color. 

\vspace{6pt} \noindent \textbf{Problem Definition.}~~
Having established the necessary preliminaries and notations, we formally define the problem addressed in our research. We consider an input triangular mesh \(\mathcal{M}\) and a textual prompt \(p\). The objective is to generate a UV texture \(\mathbf{y}\) that aligns with the semantics of \(p\) using a 2D image diffusion model and differentiable rasterization from viewpoints \(\{C_n\}\). Our goal is to ensure that the resulting images, obtained by mapping the texture \(I\) from the UV space to the intrinsic surface space of the mesh \(\mathcal{M}\), not only exhibit high perceptual quality and consistency with \(p\), but also maintain multiview consistency. It is essential for this consistency to hold even under novel view conditions, extending beyond the optimization of any pair of views from the set \(\{C_n\}\).

\section{Method} % >> [Xing] Done.
The proposed TexRO generates multi-view consistent and high-quality textures for 3D objects by optimizing its UV texture. TexRO's pipeline consists of two stages. In the first stage, we propose an effective viewpoint selection strategy that finds the smallest set of viewpoints covering a mesh's faces. In addition, we synthesize multi-view images for the given mesh and project the colors to the UV texture map to establish an initial status of the UV texture for later optimization. In the second stage, we recursively optimize the UV texture with an increasing resolution in RGB space. Specifically, we first optimize the UV texture using the proposed adaptive denoising strategy under the selected views at a low resolution, then we upsample the UV texture map and conduct the same optimization to achieve a better quality of textures at a higher resolution. We repeat this recursive process $N$ times. This design is made based on our insight that optimizing a UV texture with nearest neighbor sampling at a low resolution facilitates generating consistent content while doing so at a high resolution produces details. Our recursive optimization is thus capable of generating consistent and detailed textures. It is noteworthy that the proposed recursive optimization operates in RGB space, which ensures an accurate UV mapping during the optimization.
% unlike the previous approach, for instance, \cite{cao2023texfusion}, that optimizes latent features. Our experimental analysis validates our superiority over theirs. 
In the following sections, we introduce the proposed modules in detail. We first introduce the proposed view selection strategy in Sec.~\ref{method:view_selection}, then we introduce the proposed recursive 
optimization pipeline in Sec.~\ref{method:recursiveoptimization}. % << [Xing] Done.

\begin{figure*}[t]
\centering
\includegraphics[width=\linewidth]{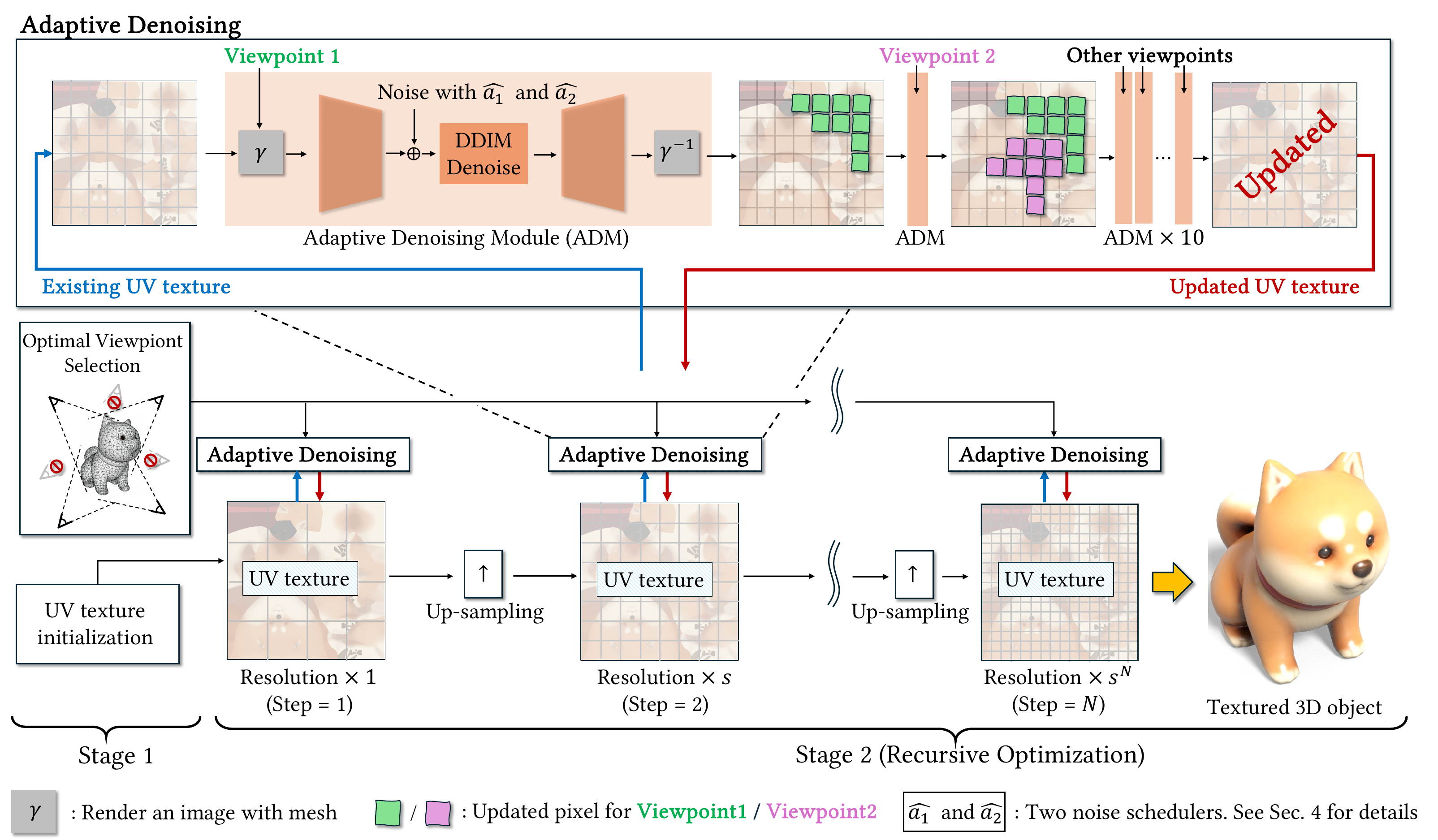}
\caption{The outline of the proposed TexRO is illustrated in the figure. It has two stages. In stage 1, it produces an optimal set of viewpoints and generates an initial UV texture for later optimization. In stage 2, it conducts the recursive optimization that optimizes the UV texture at increasing resolutions. The details of the proposed adaptive denoising strategy is illustrated at the top of the figure. $\widehat{\alpha_{1}}$ and $\widehat{\alpha_{1}}$ are two noise schedulers. }
\label{fig:pipeline}
\end{figure*}

\subsection{Optimal Viewpoints Selection} % >> [Xing] Done. 
\label{method:view_selection}
It is known that a crucial factor influencing the quality of generation is the selection of viewpoints. Previous studies such as \cite{cao2023texfusion, texturepaper} employed a predefined set of candidate viewpoints; Text2Tex~\cite{chen2023text2tex} attempted to optimize the view set but still rely on a relatively small number ($=36$) of candidates accompanied by dynamic gain computation. In this paper, we propose a viewpoints selection strategy that finds the smallest set of views covering all the faces of a mesh. This guarantees the completeness of a generated result.
Although a naive approach involves densely surrounding an object with cameras, this strategy significantly increases the optimization time. The proposed strategy formulates the goal as a \emph{Set-Cover Problem (SCP)}~\cite{cormen2022introduction} and solves it efficiently using a heuristic greedy strategy and further relaxing the 0-1 integer programming problem of minimal coverage to a general optimization problem of maximizing the sum of areas of all covered faces. In the supplementary material, we provide a detailed explanation of the algorithm and its implementation details. Table~\ref{tab:ablation_study_view_sel} validates the effectiveness of the proposed viewpoints selection method. 

\begin{figure}[t]
    \centering
    \includegraphics[width=0.95\linewidth]{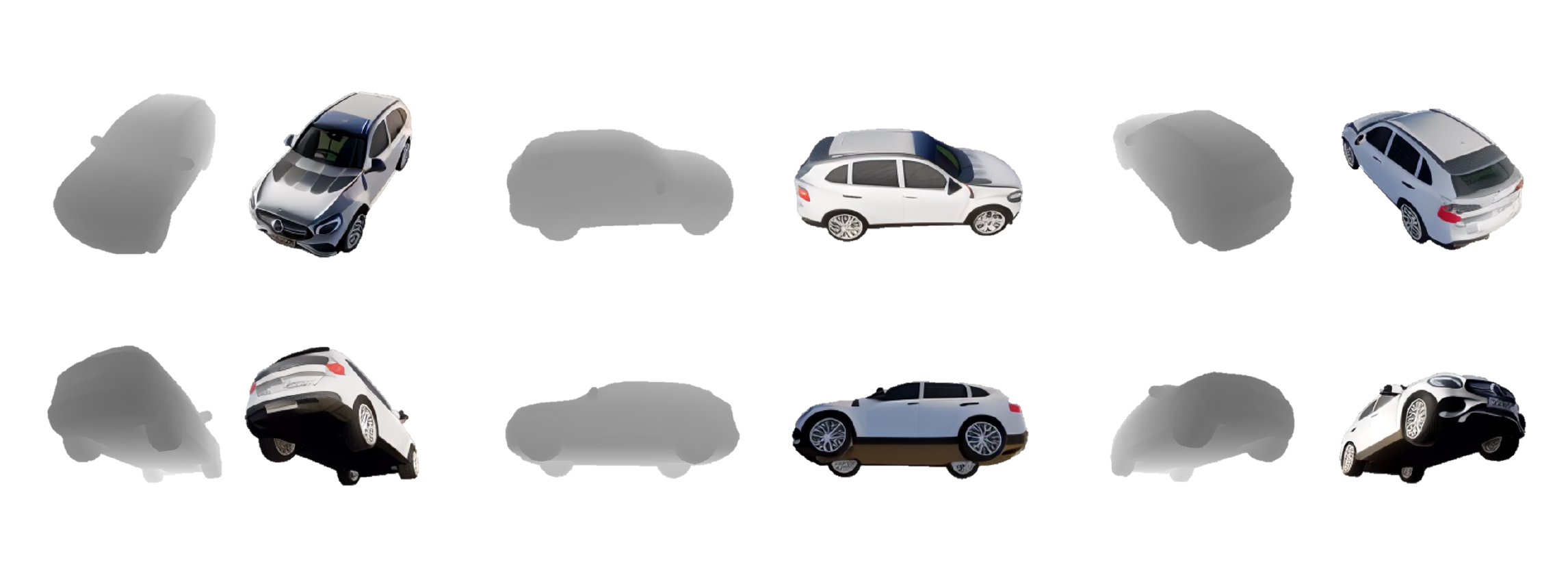}
    \caption{We employ depth-controlled Zero123Plus \cite{shi2023zero123plus} for our UV texture map initialization. The prompt for generating this example is “a Mercedes-Benz car”.}
    \label{fig:zero123results}
\end{figure}

% \begin{figure}[t]
%     \centering
%     \includegraphics[width=0.95\linewidth]{figs/denoise_process_crop.pdf}
%     \caption{\liuxing{TO BE REVISED}During the co-denoising from the optimized view set, our method progressively improves the quality of texture with regard to overall quality, detail preservation, and visual consistency. Prompt: \textit{`A photo of Napoleon Bonaparte'}.}
%     \label{fig:denoise_process}
% \end{figure}

\subsection{Recursive Optimization of UV Texture}
\label{method:recursiveoptimization}
% >>[Xing] Done.
{\bf Overview.}~ The proposed recursive optimization operates in the RGB space. It synthesizes realistic (consistent\&high visual quality) textures from a prompt by optimizing the UV texture at increasing resolutions. As illustrated in Fig.~\ref{fig:pipeline}, our optimization pipeline consists of $N$ steps. In a step, it optimizes the UV texture using the proposed adaptive denoising strategy and updates the UV texture with the optimized result. Then, it upsamples the UV texture by a factor of  $\mathbf{s}$ ($=1.5$) and optimizes it at the increased resolution using the same denoising strategy in the following step. The optimizations conducted at increasing resolutions enable it to progressively generate textural details. 
% that improve the visual quality of the generated texture. 
% We explain why this is made sure in the following paragraph. 
Our pipeline recursively conducts the explained processes for $N$ steps to generate final results.  % << [Xing] Done.

% >> [Xing] Done.
\noindent {\bf Initialization.}~
The quality of an initial UV texture affects the outcome of the final texture generation, as the proposed adaptive denoising module creates new textures based on existing ones (we introduce this in the following paragraph). We have empirically investigated that the depth-driven Zero123Plus\cite{shi2023zero123plus} serves as the most appropriate model for our UV texture initialization. This model takes an image rendered from a viewpoint as input and outputs a generated image guided by a depth map that was rendered from another point of view. We employ this model to generate six multi-view images for a mesh and project the colors from these views onto the UV texture map for initialization. Fig.~\ref{fig:zero123results} showcases a set of generated results. 
% << [Xing] Done.

% ~~ Our algorithm necessitates initialization from a coarse texture. Therefore, we require a rapid and effective method for generating coarse textures. Two approaches can serve as our initialization. The first approach involves stitching together four viewpoints onto the same grid and merging the depths into the same grid as controls, thus ensuring a certain level of consistency among multiple viewpoints. Moreover, the current multi-view diffusion \cite{shi2023zero123plus}, fine-tuned using Objaverse data, combined with operations such as reference attention, constraints consistent viewpoint generation. Zero123plus offers a superior initialization outcome. Specifically, we enforce generation across six viewpoints by rendering corresponding depths from fixed viewpoints and constraining the generation based on these depths, results as \ref{fig:zero123results}  shows.

\begin{figure}
    \centering
    \includegraphics[width=\linewidth]{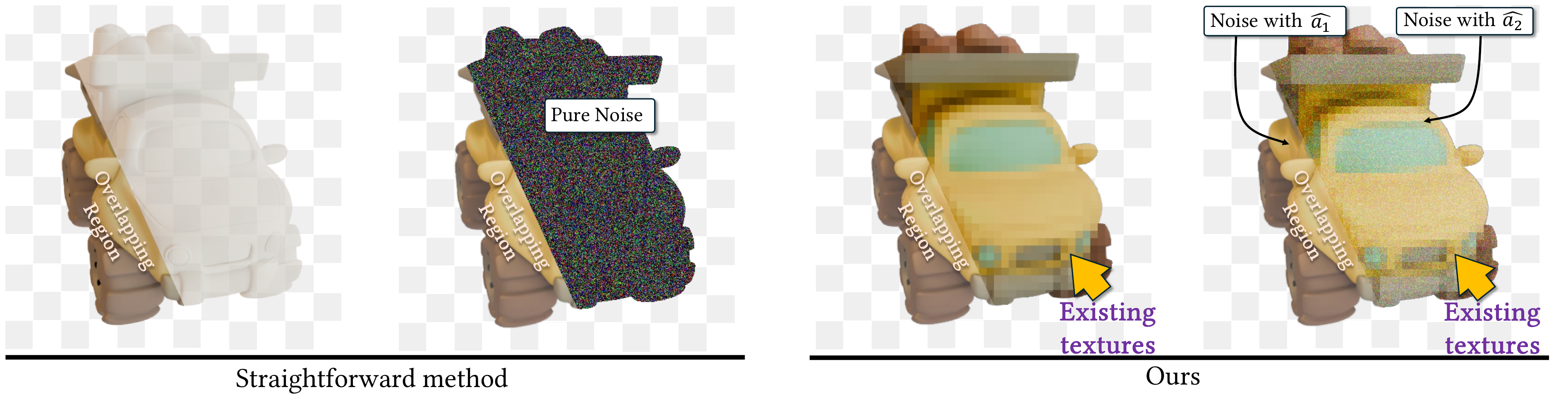}
    \caption{The key difference between the proposed adaptive denoising strategy and the straightforward method used in the previous studies \cite{chen2023text2tex, texturepaper}. In contrast to the straightforward that generates new textures from pure noise using an image inpainting diffusion model, ours injects noises to refine existing textures. We introduce how $\alpha_{t1}$ and $\alpha_{t2}$ are computed in Sec.~\ref{method:recursiveoptimization}'s adaptive denoising. }
    \label{fig:interlace}
\end{figure}

\begin{figure}[!t]
    \centering
    \captionsetup{type=figure}
    \includegraphics[width=0.95\linewidth]{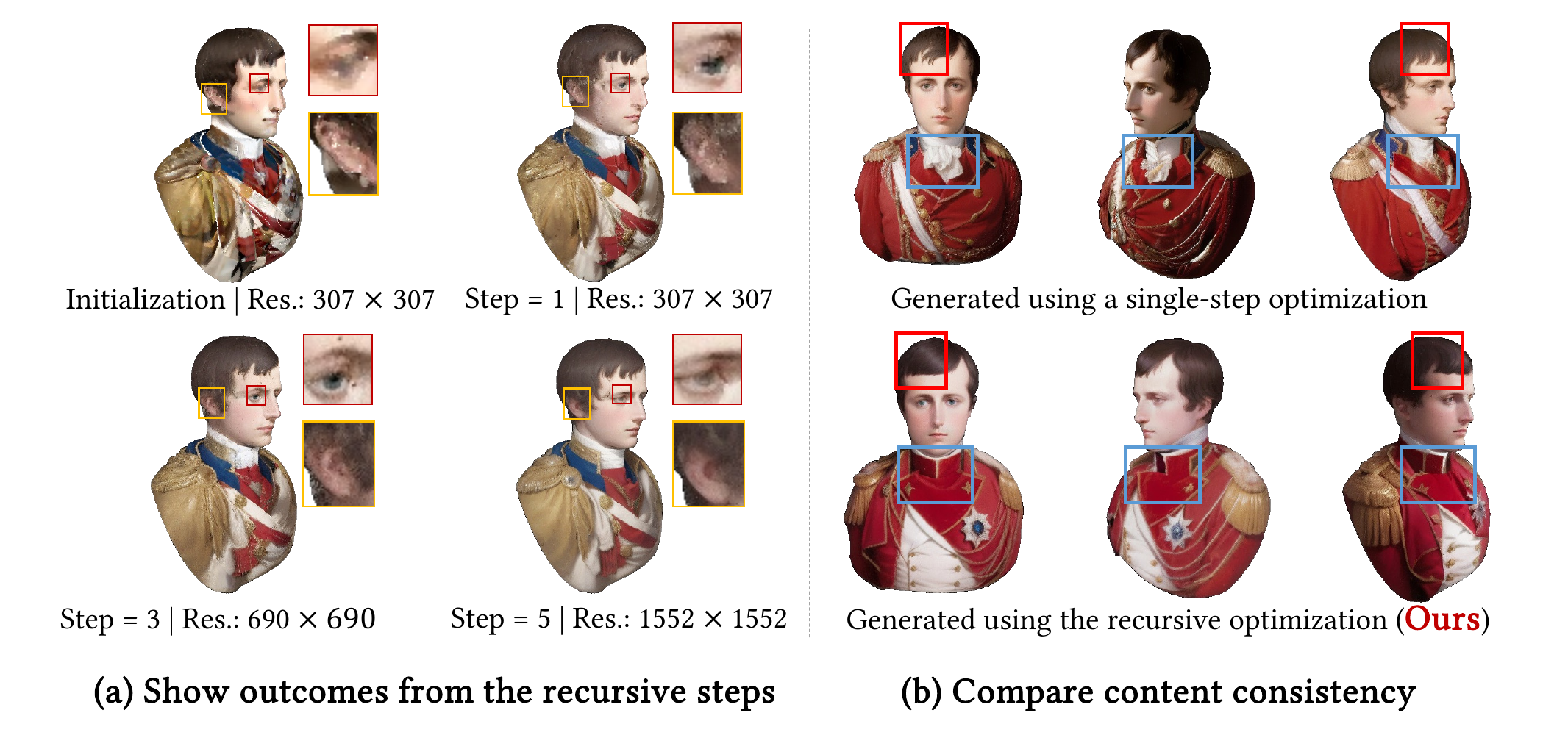}
    \captionof{figure}{(a) showcases the outcome of each recursive step in Stage-2, where “initialization” refers to the initial UV texture produced in Stage-1. (b) compares the results from the proposed recursive optimization with those from a single-step optimization performed at the highest resolution ($1552 \times 1552$), defined as the resolution applied in Step 5 of our method.}
    % The left result show that when only use single resolution texture. It will produce some unconsistency area as red} 
    \label{fig:multires}
\end{figure}

% >> [Xing] Done.
\noindent {\bf Adaptive denoising.}~ We have illustrated the outline of the proposed adaptive denoising operation at the top of Fig.~\ref{fig:pipeline}, it takes the UV texture and the selected view set as input and incrementally synthesizes textures from all the viewpoints. The key insight of the adaptive denoising strategy lies in re-using the existing textures generated from the last step (or made by the UV texture initialization when $Step = 1$) for generating new textures in the current step. As illustrated in Fig.~\ref{fig:interlace}, unlike the straightforward way \cite{chen2023text2tex, texturepaper} that generates new textures from pure noise, we inject scheduled noises into the existing textures and refine them by an image-to-image diffusion model. We compute the noises for the overlapping region and the non-overlapping region with different noise schedulers denoted as $\widehat{\alpha_{1}}$ and $\widehat{\alpha_{2}}$,  using
\begin{equation}
\begin{aligned}
\widehat{\alpha_{1}} = (\sqrt{\frac{\overline{\alpha_{t_{n}}}}{\overline{\alpha_{t_1}}}} \vz_{n, i} + \sqrt{1 - \frac{\overline{\alpha_{t_{n}}}}{\overline{\alpha_{t_1}}}} \boldsymbol{\epsilon} ) ,
\end{aligned}
\end{equation}
\begin{equation}
\begin{aligned}
\widehat{\alpha_{2}} = (\sqrt{\frac{\overline{\alpha_{t_{n}}}}{\overline{\alpha_{t_2}}}} \vz_{n, i} + \sqrt{1 - \frac{\overline{\alpha_{t_{n}}}}{\overline{\alpha_{t_2}}}} \boldsymbol{\epsilon} ) ,
\end{aligned}
\end{equation}
% -------------- Uncomment it if it is needed.
% \begin{equation}
%     \vz_{n, i} = \phi \widehat{\alpha_{1}} + (1 - \phi)  \widehat{\alpha_{2}} 
% \end{equation}
% ----------------
where $\overline{\alpha_{t_{x}}} = \prod_{u=1}^{u=t_{x}} \alpha_{u}$, $t_{n} = 10$, $t_{1} = 2$, $t_{2} = t_{n} - 1$. We have practically observed that $t_{1}$ has merely a light impact on the generation quality. We discuss the options for $t_{n}$ in Sec.~\ref{sec:exp:ablation}. $\vz_{n,i}$ means the encoded image corresponding to viewpoint $i$ and step $n$. $\epsilon$ is sampled from a normal distribution.
\vspace{3pt}\noindent {\bf Multi-resolution UV texture.}~~ We have claimed that the proposed recursive optimization approach can generate realistic textures. We explain how this is supported by the proposed multi-resolution UV texture design.
When generating textures with a low-resolution UV texture, the algorithm can retrieve more pixels from the UV texture for the overlapping region (see the evidence in supplementary material) that guides new texture generation in the adjacent viewpoints. This leads to better content consistency compared with the case where the overlapping region is small. Our design starts with a low-resolution UV texture based on this insight. To address the issue that a low-resolution UV texture can cause blocking artifacts in the generated result, we progressively increase the UV texture's resolution in the later steps (as shown in Fig.~\ref{fig:pipeline}). Although such a manner will decrease the number of pixels included in each overlapping region in a later step, content consistency will be maintained because the existing textures used in a later step are already formed with consistent structures. Fig.~\ref{fig:multires} demonstrates the correctness and effectiveness of the proposed multi-resolution UV texture design. Fig.~\ref{fig:multires}(a) showcases that consistent and high-quality textures are generated with increasing resolutions as we claimed above; Fig.~\ref{fig:multires}(b) demonstrates that the proposed approach which involves multi-resolution UV texture, generates more consistent results than a single-step optimization conducted at the highest resolution. 

\section{Experiments}

\begin{figure*}[t]
\centering
\includegraphics[width=\linewidth]{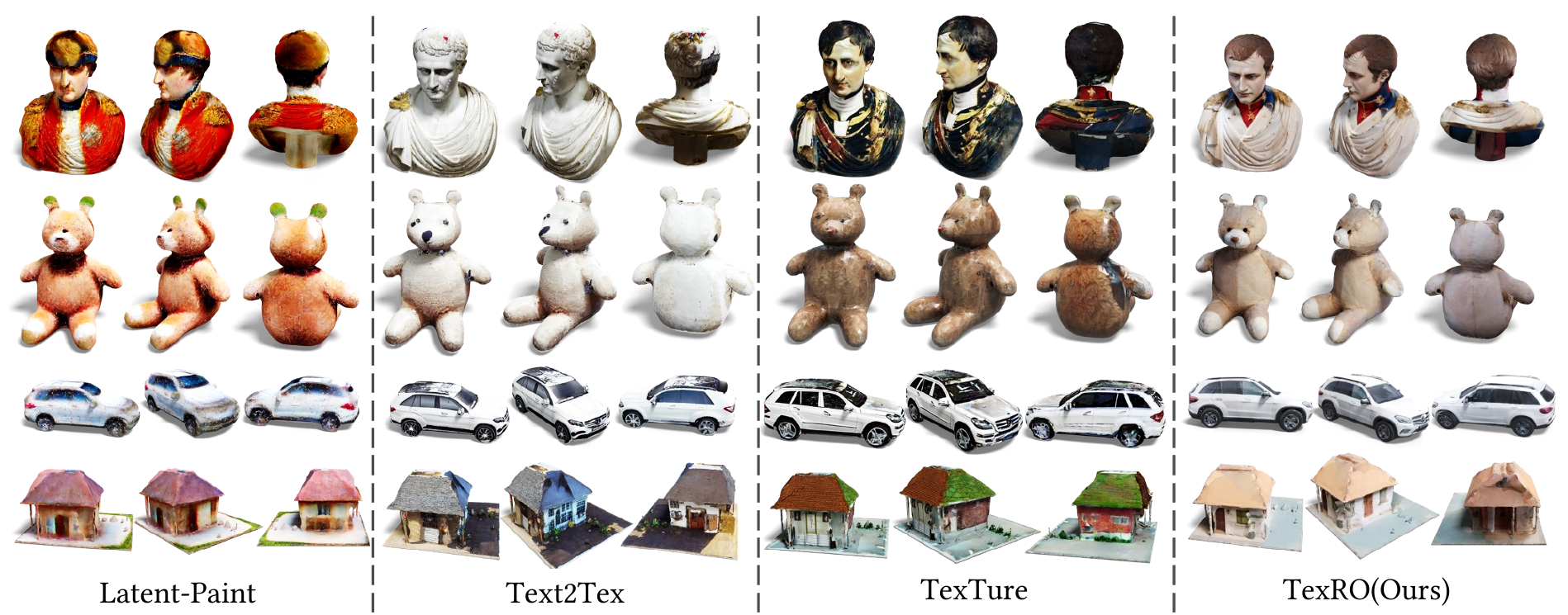}
\caption{Qualitative comparison of the generated textures among Latent-Paint~\cite{metzer2023latent}, Text2Tex~\cite{chen2023text2tex}, TEXTure~\cite{texturepaper} and our proposed TexRO.}
\label{fig:Comparison}
\end{figure*}

\subsection{Implementation Details}
\label{5.1}
We describe the implementation details. In the optimal view selection module belonging to Stage 1, we start finding the optimal set from a collection of $8192$ viewpoint candidates. To produce an initial UV texture, we employ a depth-driven Zero123Plus and six camera poses for synthesizing views. The camera poses are defined in the spherical coordinate system. Specifically, their azimuth angles are $[30, 90, 150, 210, 270, 330]$, and their elevation angles are $[60, 110, 60, 110, 60, 110]$. For the proposed recursive optimization in Stage 2, we set $N = 5$, indicating that the resolutions of UV texture are $307$, $460$, $690$, $1035$, and $1552$. We employ a depth-driven ControlNet \cite{zhang2023adding} (finetuned from Stable Diffusion v1.5) that performs image restoration to synthesize textures in the proposed adaptive denoising module. We compute $t_{n}$ for $\alpha_{t_{n}}$ for a step $n$ using
\begin{equation}
t_{n} = max(10 - s \cdot n , 5)
\end{equation}
where $s = 2.5$ in our experiment.
%we intend to XXX, as introduced in \, cite{}.}
% Our proposed method, TexRO, is designed to support text-condition modes for generating intricate textures. Our method is divided into three steps. In the first step, we fully use the existing multiview diffusion capabilities and use \cite{shi2023zero123plus} as our initialization. \cite{shi2023zero123plus} is an image condition diffusion model. To obtain the corresponding reference for the image, we use the text-to-image diffusion model and controlnet to generate the corresponding image.
% Our proposed TexRO is designed to support text-conditional modes for generating intricate textures. The sole difference between these two modes lies in the method of texture initialization at the onset of our optimization.
% The image-conditional approach is relatively straightforward, as the model presented in~\cite{shi2023zero123plus} already provides us with the ability to generate multiview images from an input image. In contrast, for the text-conditional approach, we utilize ControlNet in conjunction with depth and a textual prompt to generate an initial view. Following this, we apply the image-conditional pipeline based on the first view generated.
% We use a relatively dense selection of $K=8,192$ in all our experiments. 
We implemented our method using Python and C++ in a hybrid fashion. We employed PyTorch~\cite{pytorch} for implementing most computations. We employed Kaolin~\cite{KaolinLibrary} for differentiable rendering and texture projection. Our method is capable of generating realistic textures around \textit{only 1 minute}.
% using an NVIDIA A100 with 40GB VRAM \chenming{Suggest not mentioning A100 here}.

\subsection{Experiment Setup}

\noindent \textbf{Benchmark Dataset.}~
We employ the dataset (namely, \texttt{Text2Tex-Data}) provided in Text2Tex \cite{chen2023text2tex} for evaluation. 
% This dataset includes 800K+ 3D objects obtained from the Objaverse dataset.
In the \texttt{Text2Tex-Data}, each sample contains a mesh with textures and a brief caption describing the 3D object (using all categories as captions for objects). We compare the proposed approach and the baseline approaches using the \texttt{Text2Tex-Data}. We summarize the evaluation results in Table~\ref{tab:results}. In addition, we investigate how well the text-to-texture generation methods, including ours and the baselines, can perform when text guidance is detailed. As mentioned, the \texttt{Text2Tex-Data} only provided a brief caption for each sample. We thus bring high-quality captions provided with Cap3D \cite{luo2024scalable} into our experiment. The Cap3D's caption of an object presents more details, such as ``A 3D model of a black and white microwave oven with a power cord'' about its texture compared to the \texttt{Text2Tex-Data}'s. We reform a new evaluation dataset (denoted by \texttt{Text2Tex-Cap3d}) by taking the intersection of the \texttt{Text2Tex-Data} and the Cap3D dataset. This dataset consists of $310$ samples, each consisting of a high-quality mesh with textures and a detailed caption. We conduct the same performance comparison on the \texttt{Text2Tex-Cap3d} dataset and summarize the results in Table~\ref{tab:ablation_study_view_sel}.
Some objaverse data visual results can be found in the Fig.~\ref{fig:more3}.

\noindent \textbf{Baselines.}~
We benchmark our method against the following leading-edge text-driven texture synthesis methods: Text2Mesh~\cite{michel2022text2mesh}, CLIPMesh~\cite{mohammad2022clip}, Latent-Paint ~\cite{metzer2023latent}, Text2Tex~\cite{chen2023text2tex} and TEXTure~\cite{texturepaper}. We omitted TexFusion~\cite{cao2023texfusion} due to its closed source. 

\vspace{2pt} \noindent \textbf{Evaluation Metrics.}
We employ Fr\'{e}chet Inception Distance (FID)~\cite{heusel2017gans} and Kernel Inception Distance (KID)~\cite{sutherland2018demystifying} for evaluating the proposed approach and the baselines. For each test sample, we first synthesize textures using these approaches using the identical mesh and caption. Then, we randomly select 20 camera poses looking toward the object and render images with the synthesized textures. The groundtruth image set needed for computing FID and KID is rendered using the original texture provided with the dataset and the same 20 camera poses. We use a resolution of $512 \times 512$ for all the renderings in our evaluation.
% The quality and diversity of the generated textures are gauged using standard metrics for generative models, specifically, the Fr\'{e}chet Inception Distance (FID)~\cite{heusel2017gans} and Kernel Inception Distance (KID)~\cite{sutherland2018demystifying}. The image distribution used for these metrics includes renders of each mesh coupled with the generated textures, viewed from 20 random viewpoints, all at a resolution of $512\times  512$.

\vspace{2pt} \noindent \textbf{Runtime.}~~ One of the strengths of the proposed approach is its ability to generate textures rapidly. It can complete a generation within one minute with a single NVIDIA A100 GPU with 40G VRAM. To demonstrate our superiority, we provide a computation time comparison in Table~\ref{tab:results}. The numbers in the table validated the effectiveness of our approach.
% Our TexRO takes roughly only one minute with a single GPU to generate a texture given an input 3D mesh.  Statistics on the comparison with other approaches refer to the ``Runtime'' column of Tab.~\ref{tab:results}. 

% \input{tabs/tab6_time_comparision}
\definecolor{tabfirst}{rgb}{1, 0.7, 0.7} % red
\definecolor{tabsecond}{rgb}{1, 0.85, 0.7} % orange
\definecolor{tabthird}{rgb}{1, 1, 0.7} % yellow

\begin{table}[t]
    \centering
    \renewcommand\arraystretch{1.7}
    \tabcolsep=0.6cm
    \fontsize{7pt}{7.8pt}\selectfont
    \caption{Evaluation results on the commonly used subset of Objaverse, \textit{i.e.,} \texttt{Text2Tex-Data}, for 3D texture generation, we add TexFusion results}
    \begin{tabular}{l|ccc}
    Methods & FID $\downarrow$ & KID ($\times 10^{-3})$ $\downarrow$ & Runtime \\ \hline

Text2Mesh~\cite{michel2022text2mesh} &                      45.38 &                      10.40 &  \cellcolor{tabthird}10 min. \\
CLIPMesh~\cite{mohammad2022clip}     &                      43.25 &                      12.52 &                      50 min. \\
Latent-Paint~\cite{metzer2023latent} &                      43.87 &                      11.43 &                      22 min. \\
TEXTure~\cite{texturepaper}          &  \cellcolor{tabthird}39.09 &  \cellcolor{tabthird}9.97 & \cellcolor{tabsecond}5 min. \\
Text2Tex~\cite{chen2023text2tex}     & \cellcolor{tabsecond}35.68 & \cellcolor{tabsecond}7.74 &                      15 min. \\

% TexFusion~\cite{cao2023texfusion}     & \cellcolor{tabsecond}37.91 & \cellcolor{tabsecond}7.40 &                      6.2 min. \\
\hline
TexRO(Ours)                        &  \cellcolor{tabfirst}33.83 &  \cellcolor{tabfirst}5.77 &  \cellcolor{tabfirst}1 min.
    \end{tabular}
    \label{tab:results}
\end{table}

% \begin{table}
%     \centering
%     \caption{Evaluation results on the commonly used subset of Objaverse, \textit{i.e.,} \texttt{Objaverse-Tex}, for 3D texture generation.}
%    \resizebox{\columnwidth}{!}{
%     \begin{tabular}{l|ccccc}
%     Method & Text2Mesh~\cite{michel2022text2mesh} & CLIPMesh~\cite{mohammad2022clip} & Latent-Paint~\cite{metzer2023latent} & TEXTure~\cite{texturepaper} & Text2Tex~\cite{chen2023text2tex} & TexRO(Ours)   \\ \hline

% % Text2Mesh~\cite{michel2022text2mesh} &                      45.38 &                      10.40 &  \cellcolor{tabthird}10 min. \\
% % CLIPMesh~\cite{mohammad2022clip}     &                      43.25 &                      12.52 &                      50 min. \\
% % Latent-Paint~\cite{metzer2023latent} &                      43.87 &                      11.43 &                      22 min. \\
% % TEXTure~\cite{texturepaper}          &  \cellcolor{tabthird}39.09 &  \cellcolor{tabthird}9.97 & \cellcolor{tabsecond}5 min. \\
% % Text2Tex~\cite{chen2023text2tex}     & \cellcolor{tabsecond}35.68 & \cellcolor{tabsecond}7.74 &                      15 min. \\
% % \hline
% % TexRO(Ours)                        &  \cellcolor{tabfirst}33.83 &  \cellcolor{tabfirst}5.77 &  \cellcolor{tabfirst}1 min.
%     \end{tabular}
%     }
%     \label{tab:results}
% \end{table}
\subsection{Quantitative Comparison}
The quantitative results computed on \texttt{Text2Tex-Data} are provided in Table~\ref{tab:results}. The result values for the baseline methods are borrowed from \cite{chen2023text2tex}. The table shows that the proposed approach outperforms the baseline methods significantly. Our approach also achieves the shortest computation time. In Table~\ref{tab:ablation_study_view_sel}, we provide the quantitative results computed on the \texttt{Text2Tex-Cap3d} dataset. 
% It is noticed that we did not run evaluations for all the baseline methods except Text2Tex \cite{chen2023text2tex}. The reason is that the previous methods are computationally heavy, it takes around a week to finish the computation using our computing resources\chenming{I suggest removing this statement}. 
Due to the limitation of our computing resources, we only compare the proposed approach with the strongest baseline (\ie Text2Tex \cite{chen2023text2tex}) on the \texttt{Text2Tex-Cap3d} dataset to demonstrate our superior effectiveness. It is seen in Table~\ref{tab:results} that the proposed approach outperforms the baseline methods. It is noteworthy that involving detailed captions can improve the quality of texture generation, as we have stated before. See the result values of the proposed approach shown in both Table~\ref{tab:results} and  Table~\ref{tab:ablation_study_view_sel}-TexRO(\textit{w/} View Sel.).\\
To further validate the effectiveness of the proposed approach, we conducted a user study for comparison. The user study involved 20 participants who possess fundamental knowledge of 3D modeling. We randomly select 100 samples from the \texttt{Text2Tex-Data} and conduct pairwise comparisons. For each sample, we produce two images rendered using the UV texture maps generated from the proposed approach and Text2Tex \cite{chen2023text2tex} with an identical camera pose. Consequently, we asked the participants to output pairwise preference for each sample based on overall perceptual quality, which includes 1) multi-view consistency, 2) detail richness, and 3) natural color representation. The results of this user study indicate a clear preference for the proposed approach. Compared to the results of Text2Tex \cite{chen2023text2tex}, Ours received a favorable verdict, with a preference ratio of $68.4\%$ versus $31.6\%$. This reinforces the superior quality of the textures generated by our method. We conduct extensive comparisons between our proposed method and \cite{cao2023texfusion} and report the results in the supplementary material.

% \input{tabs/tab2_metric_shapenet}
% We conduct a thorough evaluation of TexRO against other similar methods, employing metrics such as FID and KID for the evaluation. Details of these experimental results can be found in Tab.~\ref{tab:results}. The results highlight the overall superiority of our proposed method, boasting a significant improvement in comparison to the others. This underscores the effectiveness of TexRO in achieving high-quality 3D texture generation.

% To further validate our findings, a user study was carried out involving 20 participants possessing fundamental knowledge of 3D modeling.  The study was designed around pairwise preferences, focusing on the overall perceptual quality based on various criteria, including 1) multiview consistency, 2) detail richness, and 3) natural color representation. The results of this user study indicate a clear preference for TexRO. Compared to the results of \cite{chen2023text2tex}, TexRO received a favorable verdict, with a preference ratio of $68.4\%$ versus $31.6\%$. This reinforces the superior quality of the textures generated by our method.

% Importantly, our proposed TexRO has proven to be particularly effective when implemented on the `car' subcategory within the ShapeNet dataset. The robust performance in this application further establishes the versatility and adaptability of our method across different contexts. Due to space constraints within this section, we have included a detailed report of these experimental results in the supplementary material provided.
% \input{tabs/tab4_na}

\subsection{Qualitative Results}
We present our qualitative results in Fig.~\ref{fig:figInconsistent} and Fig.~\ref{fig:more3}. All our renderings are conceived from novel viewpoints, showcasing the versatility and adaptability of our approach. 
% % To provide a more comprehensive understanding and evaluation of the qualitative results produced by our method, we have included circular rendered videos in our supplementary material.
In addition, more visual results are available in Fig.~\ref{fig:Comparison} to further evaluate the quality of our generated 3D textures. This visual comparison further emphasizes the level of detail and accuracy achieved by our approach. 

%Moreover, we have provided an array of qualitative comparisons in the supplementary material. This comparative framework allows for a more balanced and informed assessment of our method's performance relative to existing approaches.
% We show the generated text-conditional 3D textures with our proposed TexRO in Fig.~\ref{fig:teaser}, as stated in the caption, all of our renderings are made in novel views. We provide circular rendered videos in our suppmentary material to help better understand and evaluate the qualitative results produced by our proposed method.
% As shown in Fig.~\ref{fig:Comparison}, we provide more visual results to help evaluate the quality of our generated 3D textures. Besides, more qualititive comparisons are provided in the supplementary material, including rendered views uniformly sampled from the bounding spheres, along with the results of other baseline methods rendered at the same viewpoints.

% \input{tabs/tab6_time_comparision}

% \input{tabs/tab4_na}

\definecolor{tabfirst}{rgb}{1, 0.7, 0.7} % red
\definecolor{tabsecond}{rgb}{1, 0.85, 0.7} % orange
\definecolor{tabthird}{rgb}{1, 1, 0.7} % yellow

% \begin{figure}
% \begin{floatrow}
% \capbtabbox{%
% \resizebox{0.4\textwidth}{6mm}{
%   \begin{tabular}{l|cc}
% Method & ~FID $\downarrow$ &~~ KID ($\times 10^{-3}$) $\downarrow$\\ 
% \hline
% TexRO (w/o VOPT.) & 32.75 &  5.42 \\
% TexRO (w/  VOPT.) &  \cellcolor{tabfirst}30.33  &  \cellcolor{tabfirst}4.58 \\
% \end{tabular}}
% }{%
%   \caption{The results are computed on \texttt{Text2Tex-Cap3d}. In the left table, TexRO (w/o View Sel.) and TexRO (w/ View Sel.) mean the texture generations without and with the proposed view selection strategy. In the case of TexRO (w/o View Sel.), the views are uniformly distributed. }%
% }

% \begin{figure}
%     \includegraphics[width=0.4\textwidth]{figs/ablation_study_denoise_step.png}
%      \captionof{figure}{The impact of distinct denoising step counts on FID and KID.} 
% \end{figure}
% \end{floatrow}
% \end{figure}

\begin{figure}[t]
  \begin{minipage}[b]{.45\linewidth}
    \centering
    \label{fig:ablation_study_denoise_step}
    \includegraphics[width=1.0\linewidth]{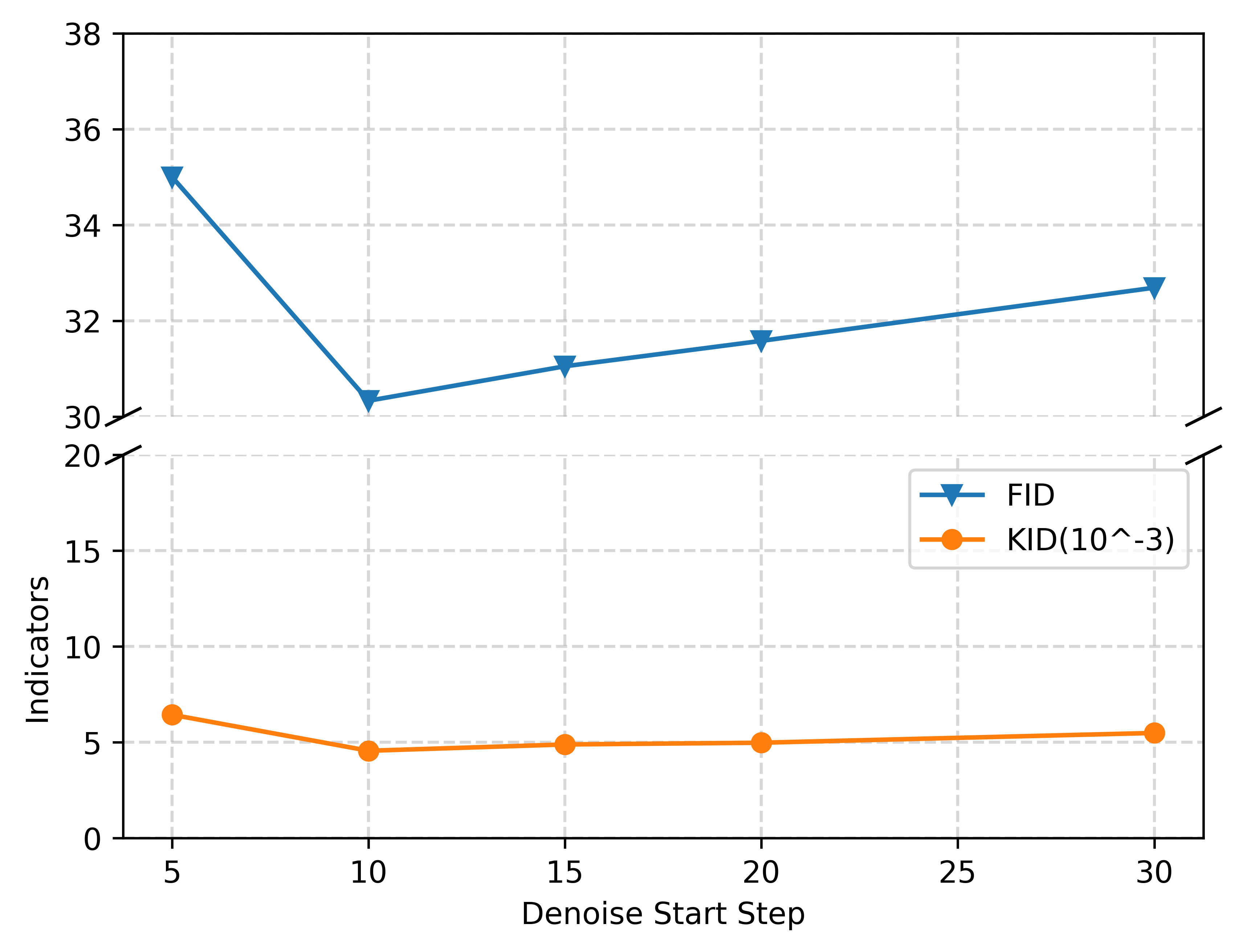}
    \captionof{figure}{The impact of distinct denoising step counts on FID and KID.}% \caption{Figure caption}
    \label{fig:ablation_study_denoise_step}
  \end{minipage}\hfill
  \begin{minipage}[b]{.5\linewidth}
  \centering
  \captionof{table}{The results are computed on \texttt{Text2Tex-Cap3d}. In the left table, TexRO (\textit{w/o} View Sel.) and TexRO (\textit{w/} View Sel.) mean the texture generations without and with the proposed viewpoint selection strategy. In the case of TexRO (\textit{w/o} View Sel.), the viewpoints are uniformly distributed. 
  \vspace{2em}}
  \resizebox{\columnwidth}{!}{
  \renewcommand\arraystretch{1.5}
  \begin{tabular}{l|cc}
    Method & ~FID $\downarrow$ &~~ KID ($\times 10^{-3}$) $\downarrow$\\ 
    \hline
    Text2Tex~\cite{chen2023text2tex} & \cellcolor{tabthird}34.14 & \cellcolor{tabthird}5.94 \\
\hline 
    TexRO (\textit{w/o} View Sel.) & 32.75 &  5.42 \\
    TexRO (\textit{w/}~~View Sel.) &  \cellcolor{tabfirst}30.33  &  \cellcolor{tabfirst}4.58 \\
\end{tabular}
    }
    \label{tab:ablation_study_view_sel}
  \end{minipage}
\end{figure}

\subsection{Ablation Study}
\label{sec:exp:ablation}
% \vspace{3pt} \noindent \textbf{Model choice on texture initialization.}~~Our view-based texture initialization module only needs coarse initialization as a guide to later fine granular optimization. To demonstrate the capability of taking different levels of coarseness as input (Fig.~\ref{fig:diff_init}), we use basic ControlNet with four views and Zero-123, and the results are shown in the Fig.~1 of the supplementary material.

% we use ControlNet s input is 2x2 grid conditions, We spliced the four condition perspectives into a 2x2 grid, and used controlnet to simultaneously generate images under the four conditions. Generating images from different perspectives at the same time can ensure a certain consistency, this model is only with basic generalization and multiview consistency.
% We evaluate the results using the initialized textures from such model and ~\cite{shi2023zero123plus} using the metrics, the evaluated results are shown in Tab.~\ref{xx}, which verifies there is no significant gap between those two choices of the model used to initialize texture, demonstrating the robustness of our proposed optimization method. We also provide a few visual comparisons in the supplementary material.

\vspace{3pt} \noindent \textbf{Effectiveness of view selection.}~
We demonstrate the effectiveness of the proposed optimal view selection by conducting experiments on \texttt{Text2Tex-Cap3d}. Table~\ref{tab:ablation_study_view_sel} shows the result values computed \textit{w/} and \textit{w/o} the proposed optimal view selection method. It is seen that the proposed method effectively improves the performance of texture generation by $1.70$ and $0.56 (\times 10^{-3})$ for FID ($\downarrow$) and KID ($\downarrow$). \\ 

\vspace{3pt} \noindent \textbf{Choices of values for $\alpha_{t_{n}}$.}~
We have introduced in Sec.~\ref{method:recursiveoptimization} that we gradually decrease the noise level with increasing step $n_i$. We analyze how different starting values of noise level (\ie the noise level at $n_i = 1$) affect texture generation results. We provide quantitative results computed using different starting values of the noise level in Fig.~\ref{fig:ablation_study_denoise_step}. It is seen that $10$ gives the best performance compared to other options. \\

\vspace{1cm}
\section{Conclusion}
We introduce TexRO, a novel method for generating realistic textures on 3D geometries. The key innovations lie in the proposed recursive optimization pipeline and the optimal view selection strategy. The recursive optimization utilizes an interlaced denoising strategy that leverages the advancements in diffusion probabilistic models to synthesize textures at increasing resolutions. This facilitates it to first generate consistent structures, and then enhance details of textures. Through rigorous experimentation and user studies, TexRO has proven to significantly surpass existing methods in delivering superior texture quality, remarkable detail preservation, enhanced visual consistency, and improved runtime efficiency, applicable across various 3D models. 
\vspace{3pt}
\\
% We introduce TexRO, a novel method for generating realistic textures on 3D geometries, leveraging the advancements in diffusion probabilistic models and multiview optimization. The core of TexRO's innovation lies in its unique approach to texture generation. It starts with a coarse texture map created from a multiview diffusion model, then refined through a multiview optimization process. This process utilizes an interlaced denoising strategy and an optimized viewpoint selection, resulting in textures that are not only realistic but also consistent across multiple views. Through rigorous experimentation and user studies, TexRO has proven to significantly surpass existing methods in delivering superior texture quality, remarkable detail preservation, enhanced visual consistency, and improved runtime efficiency, applicable across various 3D models.
{\bf Limitation.~}
While the proposed TexRO demonstrates effectiveness in generating realistic textures for a wide range of 3D models, it encounters two primary limitations. First, it requires the input meshes to be water-tight (2-manifolds), which may limit its applicability to industrial uses, such as repainting objects whose geometry has been reconstructed using photometric methods. Investigating this will be a priority in our future work. Secondly, TexRO occasionally fails to accurately texture areas within meshes that possess complex topologies. We believe that direct optimization on the mesh surfaces could overcome this issue. Lastly, we aim to extend TexR’s capabilities for generating physically-based rendering (PBR) materials, aligning with modern rendering engine standards.

\begin{figure*}[t!b]
\vspace{1cm}
\centering
\includegraphics[width=\linewidth]{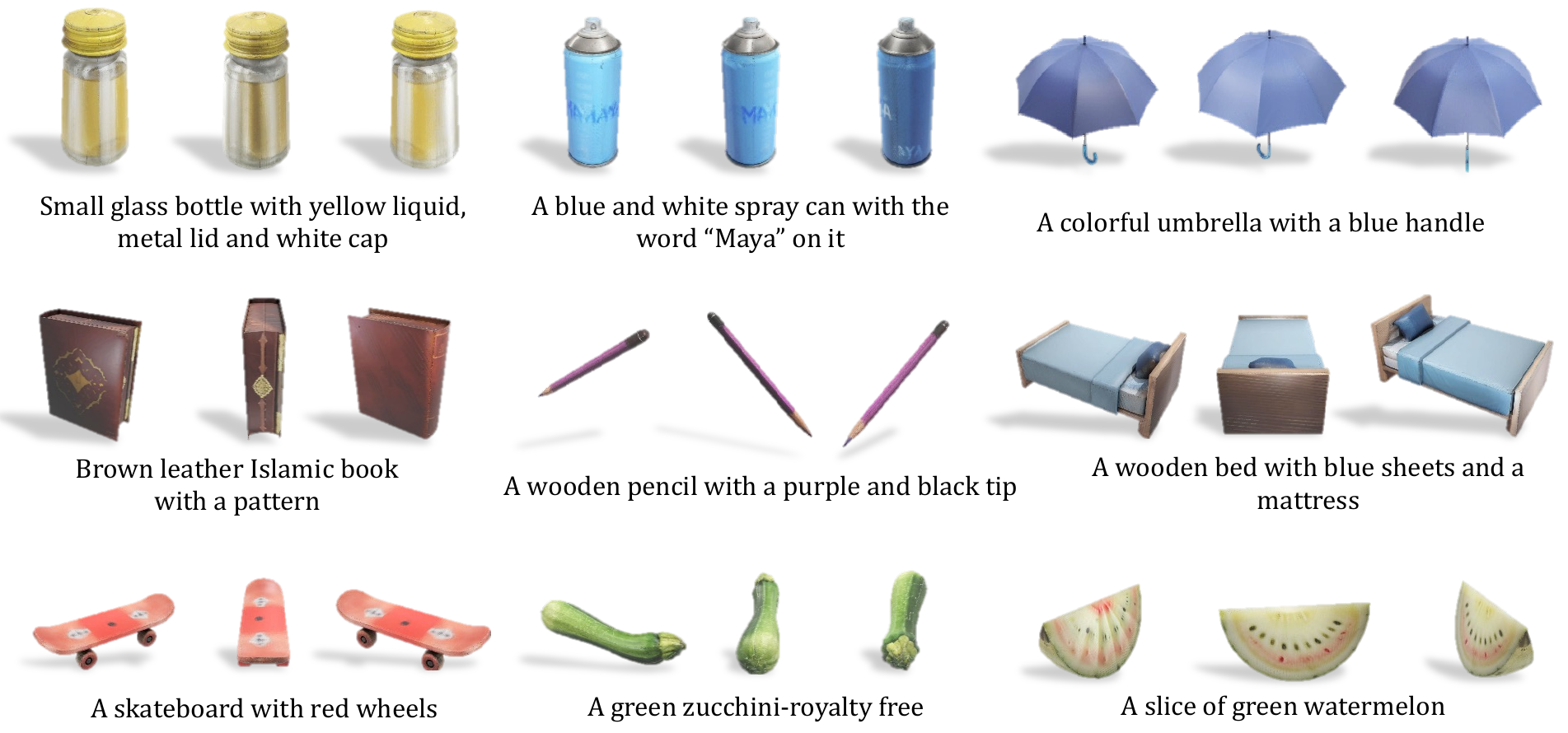}
\caption{Additional results generated by our proposed TexRO method using texts as prompts. More results are showed in our supplementary material.}
\label{fig:more3}
\end{figure*}

\clearpage
\bibliographystyle{splncs04}
\bibliography{main}

\clearpage
% \clearpage
% \setcounter{page}{1}
% \maketitlesupplementary
\setcounter{figure}{7}
\setcounter{table}{2}
% {\bf{Overlap Mask in Multi-Res UV texture}}~ As mentioned in \ref{}, different resolutions of UV texture will cause different overlaps in the view image. Fig.~\ref{} shows the different size of overlap area under three different resolutions of UV texture.
\begin{figure*}[!h]
    \centering
    \includegraphics[width=\linewidth]{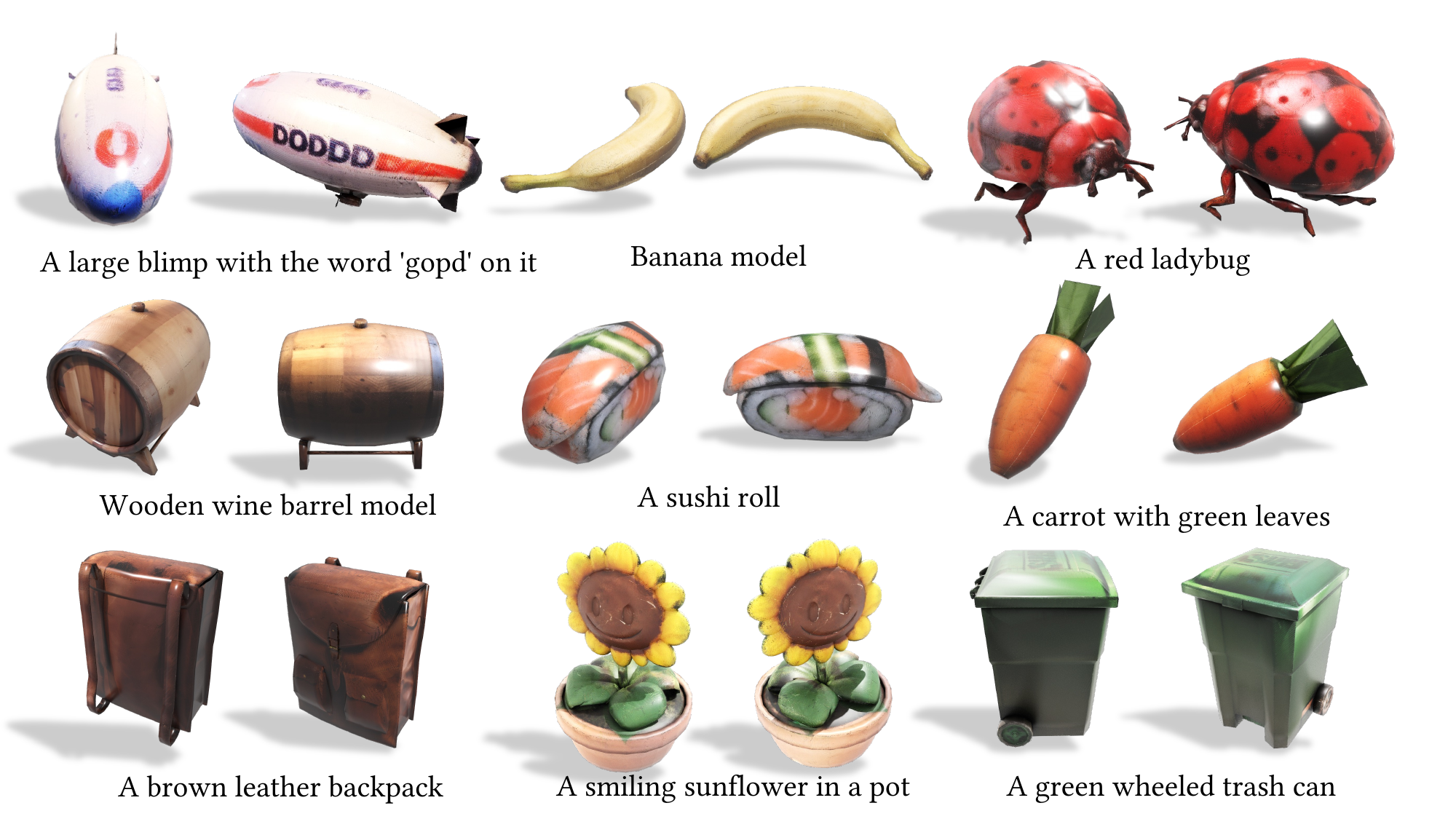}
    \caption{More results from the proposed TexRO.}
    \label{fig:supp-moreresults}
\end{figure*}

\section{Implementation Details and Additional Results}
\noindent {\bf{Prompt Augmentation}}~~ We employ prompt augmentation to improve the performance of the stable diffusion models used in our method, following [32]. We add the word ``front'' when azimuth lies in $[0^{\circ}$, $30^{\circ}]$ or $[330^{\circ}, 360^{\circ}]$ to the original prompt; add the word ``back'' when azimuth lies in $[30^{\circ}$, $150^{\circ}]$ or $[210^{\circ}, 330^{\circ}]$ to the original prompt.

% Follow the previous sds work \cite{poole2022dreamfusion}, we also use extra words in our prompt to represent the direction of every view point, since we use spherecial coordinate to represent camera pose. When azimuth is from ($0^{\circ}$ to $30^{\circ}$) or ($330^{\circ}$ to $360^{\circ}$) , we set prompt of format "{prompt}, front", meanwhile side range is ($30^{\circ}$ to $150^{\circ}$) and ($210^{\circ}$ to $330^{\circ}$). The other range we set it as "back" view.
% \\
\noindent {\bf More results from TexRO}~~ We provide more results generated using our TexRO in Fig.~\ref{fig:supp-moreresults}.

\section{Explanation of Optimal Viewpoints Selection}
We explain the details of the proposed viewpoints selection algorithm. We have explained in Sec.3-Preliminary of the main text that a 3D mesh is represented by triangle faces. The goal of the proposed viewpoints selection method is to find the smallest set of viewpoints covering all the faces of a mesh.
\begin{lemma} 
Let $U$ and $\mathcal{A}$ represent two universal sets, where $U$ is the set of all triangle faces of a mesh, and $\mathcal{A}$ is the set of all possible viewpoints that surround the mesh and are oriented towards it.
% Given two universal sets $U$ and $\mathcal{A}$, where $U$ denotes the union of triangle faces of a mesh and $\mathcal{A}$ denotes the union of possible viewpoints surrounding and looking towards the mesh. 
Let $S$ be a collection of subsets of $\mathcal{A}$, then, a set cover is a sub-collection $S'$ of $S$ such that every element in $U$ is ``covered'' (=viewed by) at least one subset in $S'$. The goal is to find the smallest possible $S'$ that covers all elements in $U$. 
\end{lemma}
\begin{theorem}
Let $U = \{u_1, u_2, \dots, u_n\}$ be the universal set with $n$ elements; let $S = \{S_1, S_2, \dots, S_m\}$ be the collection of subsets of $U$ with $m$ subsets. We define a binary variable $x_j$ for each subset $S_j$ in $S$ where:
\begin{equation}
\left\{\begin{matrix}
 x_j =& ~1 ~~~~~\text{if}~~ S_j \subset S' \\
 x_j =& 0 ~~~~~\text{otherwise.}
\end{matrix}\right. 
\end{equation}
The objective is to minimize the number of subsets selected in the set cover $S'$, which can be formulated as:
\begin{equation}
min \sum^{m}_{j=1}x_j, \quad\text{s.t.}\quad\sum_{j:u_i  
\triangleleft S_j}x_j \geq 1, \quad\forall i \in \{1,2,\dots,n\}, \forall j \in \{1,2,\dots,m\}, 
\end{equation}
where $u_i \triangleleft S_j$ means a triangle face $u_i$ is ``covered'' (=viewed) by at least one subset of viewpoints in $S'$. 
\end{theorem}
{\bf Solution}~~ we solve the aforementioned problem using a heuristic greedy strategy and further relaxing the 0-1 integer programming problem of minimal coverage to a general optimization problem of maximizing the sum of areas of all covered faces. Specifically, we commence by sampling $K$ candidate viewpoints from a spherical surface with a variable radius ranging from $1.0$ to $1.4$ (this assumes that the 3D mesh is normalized in the range of $[-1, 1]$). Following this, we assess the visibility between a viewpoint and each face of the mesh using ray-face intersection tests.  We stipulate that the included angle in this interaction should be less than $45^{\circ}$. This results in an optimization that can be embarrassingly paralleled, typically clocking in at $1$ to $2$ seconds for the majority of models with a dense selection of $K$. We use $K=8192$ for all our experiments.\\

\begin{figure}[t]
    \centering
    \includegraphics[width=\linewidth]{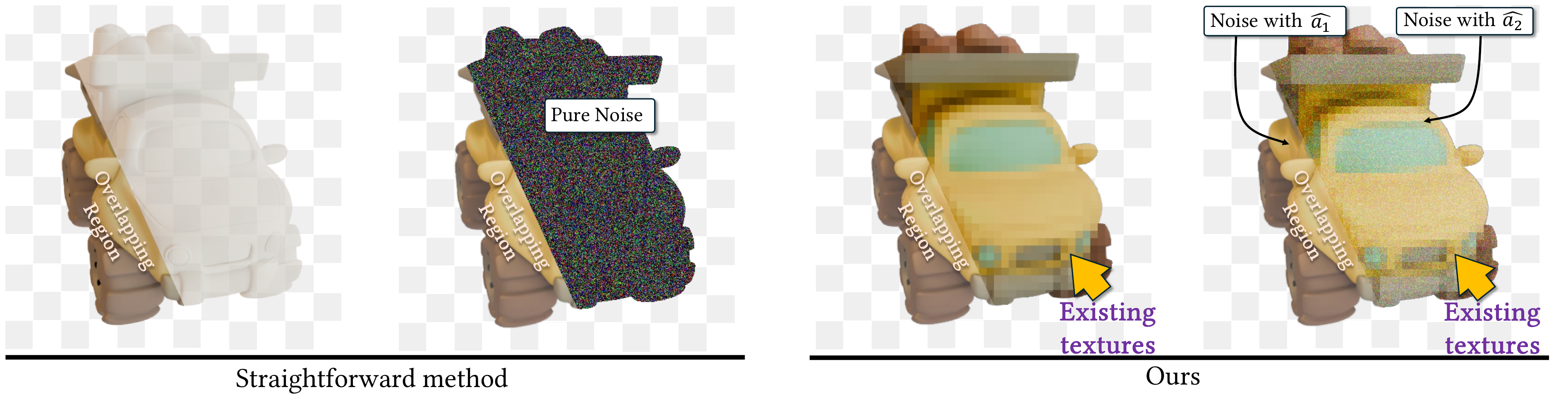}
    \caption{{\bf The duplicate of Fig.4 for ease of reference} ~~The key difference between the proposed adaptive denoising strategy and the straightforward method used in the previous studies [4, 34]. In contrast to the straightforward that generates new textures from pure noise using an image inpainting diffusion model, ours injects noises to refine existing textures. We introduce how $\alpha_{t1}$ and $\alpha_{t2}$ are computed in Sec. 4.2's adaptive denoising. }
    \label{sup_fig:adaptive_denoising}
\end{figure}

\section{Additional Explanation of Adaptive Denoising}
\begin{figure}[t]
    \centering
    \includegraphics[width=\linewidth]{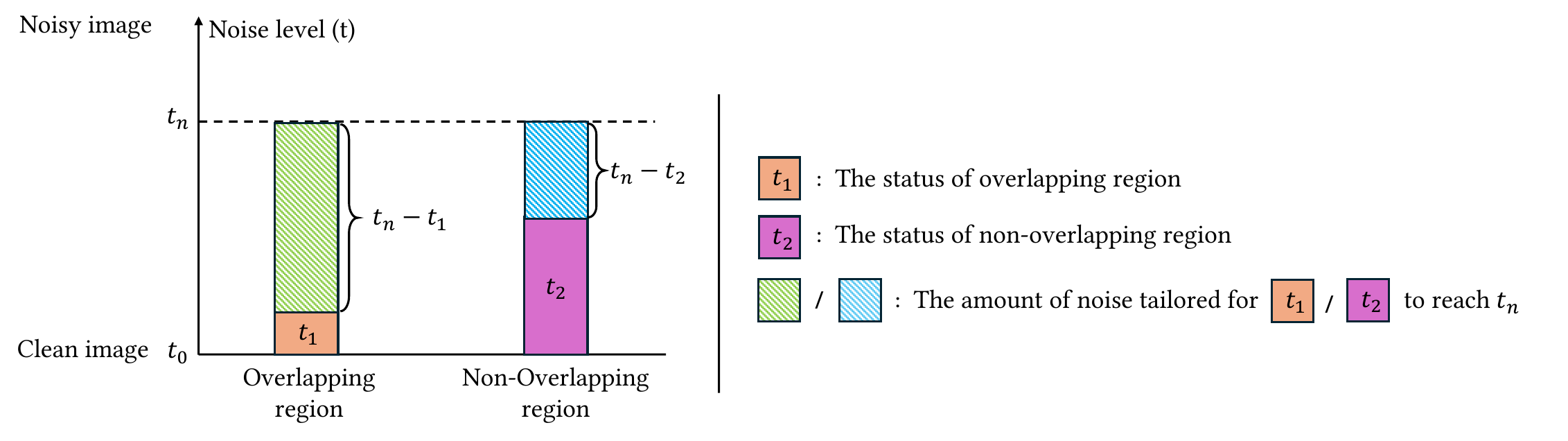}
    \caption{The illustration demonstrates the essence of the proposed adaptive denoising strategy, explaining our intention to employ two distinct noise schedulers to diffuse the pixels in the overlapping and non-overlapping regions.}
    \label{sup_fig:adp_alphas}
\end{figure}

We have illustrated the essence of our proposed adaptive denoising strategy in Fig.4 of the main text. For ease of reference, we have replicated the figure and included it here (Fig.~\ref{sup_fig:adaptive_denoising}). We explain why we employ two noise schedulers, $\widehat{\alpha_1}$ and $\widehat{\alpha_2}$, for distinct applications to the overlapping and non-overlapping regions. We present our rationale behind naming the method "adaptive denoising" to the reader. The ``Overlapping'' region contains the co-visible pixels, whose colors are the outcomes of a previous adaptive denoising process, executed from a viewpoint adjacent to the current one. As illustrated in Fig.~\ref{sup_fig:adp_alphas}, we believe that the textures in the ``Overlapping region'' are in a state (denoted as $t_1$ in the figure) closer to the clean image, in terms of diffusion denoising. Conversely, the textures in the non-Overlapping region are in a state (denoted as $t_2$ in the figure) that is further from the clean image. In this scenario, we need to apply tailored noises (represented by $t_n - t_1$, $t_n - t_2$ in the figure) across the different regions (\ie, overlapping and non-overlapping) to diffuse the pixels to a uniform level (denoted as $t_n$ in the figure). We thus employ two noise schedulers ($\widehat{\alpha_1}$ and $\widehat{\alpha_2}$) to achieve this purpose, we calculate them using equations (2) and (3) written in the main text.

\section{Comparison with Texfusion}
We conduct a comparative analysis between our TexRO and Texfusion [2]. Since Texfusion's code is not publicly available, we implemented its method ourselves, striving to replicate its original functionality accurately. We ensured the comparison was carried out in an identical experimental setting. Notably, Texfusion uses varying resolutions for UV maps for different samples, as mentioned in their paper (Sec.4.2.2). In contrast, our implementation uses an identical resolution for the UV maps for all the samples. We mimic the experimental details introduced in Texfusion's paper, we utilize 24 viewpoints in the first stage and 9 viewpoints in the second stage. We conduct the comparison on Text2Tex-Data and Text2Tex-Cap3D. We do not use the samples used in the original Texfusion's paper due to their cost. The quantitative comparison is shown in Table~\ref{tab:withtexfusion} and Table~\ref{tab:texfusion_cap3d}. It is seen that the proposed TexRO outperforms Texfusion by a good margin in terms of texture generalization quality and running time.
\definecolor{tabfirst}{rgb}{1, 0.7, 0.7} % red
\definecolor{tabsecond}{rgb}{1, 0.85, 0.7} % orange
\definecolor{tabthird}{rgb}{1, 1, 0.7} % yellow

\begin{table}[t]
    \centering
    \renewcommand\arraystretch{1.7}
    \tabcolsep=0.6cm
    \fontsize{7pt}{7.8pt}\selectfont
    \caption{Evaluation results on the commonly used subset of Objaverse, \textit{i.e.,} \texttt{Text2Tex-Data}, for 3D texture generation}
    \begin{tabular}{l|ccc}
    Methods & FID $\downarrow$ & KID ($\times 10^{-3})$ $\downarrow$ & Runtime \\ \hline

TexFusion  $\dagger$ ~[2]     & 37.91 & 7.40 &                      6.2 min. \\

\hline
TexRO(Ours)                        &  \cellcolor{tabfirst}33.83 &  \cellcolor{tabfirst}5.77 &  \cellcolor{tabfirst}1 min.
    \end{tabular}
    \label{tab:withtexfusion}

\end{table}

%-------------------------------------------------
% \begin{table}[t]
%     \centering
%     \renewcommand\arraystretch{1.7}
%     \tabcolsep=0.6cm
%     \fontsize{7pt}{7.8pt}\selectfont
%     \caption{Evaluation results on the commonly used subset of Objaverse, \textit{i.e.,} \texttt{Text2Tex-Data}, for 3D texture generation}
%     \begin{tabular}{l|ccc}
%     Methods & FID $\downarrow$ & KID ($\times 10^{-3})$ $\downarrow$ & Runtime \\ \hline

% Text2Mesh~\cite{michel2022text2mesh} &                      45.38 &                      10.40 &  \cellcolor{tabthird}10 min. \\
% CLIPMesh~\cite{mohammad2022clip}     &                      43.25 &                      12.52 &                      50 min. \\
% Latent-Paint~\cite{metzer2023latent} &                      43.87 &                      11.43 &                      22 min. \\
% TEXTure~\cite{texturepaper}          &  \cellcolor{tabthird}39.09 &  \cellcolor{tabthird}9.97 & \cellcolor{tabsecond}5 min. \\
% Text2Tex~\cite{chen2023text2tex}     & \cellcolor{tabsecond}35.68 & \cellcolor{tabsecond}7.74 &                      15 min. \\

% TexFusion~\cite{cao2023texfusion}     & \cellcolor{tabsecond}37.91 & \cellcolor{tabsecond}7.40 &                      6.2 min. \\

% \hline
% TexRO(Ours)                        &  \cellcolor{tabfirst}33.83 &  \cellcolor{tabfirst}5.77 &  \cellcolor{tabfirst}1 min.
%     \end{tabular}
%     \label{tab:withtexfusion}
% \end{table}

\definecolor{tabfirst}{rgb}{1, 0.7, 0.7} % red
\definecolor{tabsecond}{rgb}{1, 0.85, 0.7} % orange
\definecolor{tabthird}{rgb}{1, 1, 0.7} % yellow

% \begin{figure}
% \begin{floatrow}
% \capbtabbox{%
% \resizebox{0.4\textwidth}{6mm}{
%   \begin{tabular}{l|cc}
% Method & ~FID $\downarrow$ &~~ KID ($\times 10^{-3}$) $\downarrow$\\ 
% \hline
% TexRO (w/o VOPT.) & 32.75 &  5.42 \\
% TexRO (w/  VOPT.) &  \cellcolor{tabfirst}30.33  &  \cellcolor{tabfirst}4.58 \\
% \end{tabular}}
% }{%
%   \caption{The results are computed on \texttt{Text2Tex-Cap3d}. In the left table, TexRO (w/o View Sel.) and TexRO (w/ View Sel.) mean the texture generations without and with the proposed view selection strategy. In the case of TexRO (w/o View Sel.), the views are uniformly distributed. }%
% }

% \begin{figure}
%     \includegraphics[width=0.4\textwidth]{figs/ablation_study_denoise_step.png}
%      \captionof{figure}{The impact of distinct denoising step counts on FID and KID.} 
% \end{figure}
% \end{floatrow}
% \end{figure}

\begin{table}[t]
  \centering
   \renewcommand\arraystretch{1.7}
    \tabcolsep=0.7cm
    \fontsize{7pt}{7.8pt}\selectfont
  \caption{The results are computed on \texttt{Text2Tex-Cap3d}.
  \vspace{2em}}
  \resizebox{0.8\columnwidth}{!}{
  \renewcommand\arraystretch{1.5}
  \begin{tabular}{l|cc}
    Method & ~FID $\downarrow$ &~~ KID ($\times 10^{-3}$) $\downarrow$\\ 
    \hline
    % Text2Tex~\cite{chen2023text2tex} & \cellcolor{tabthird}34.14 & \cellcolor{tabthird}5.94 \\
\hline 
    TexFusion $\dagger$ [2]  & 35.35 &  5.82 \\
    TexRO (Ours)&  \cellcolor{tabfirst}30.33  &  \cellcolor{tabfirst}4.58 \\
\end{tabular}
    }
    \label{tab:texfusion_cap3d}
\end{table}

\end{document}